\documentclass[letterpaper,10pt,conference]{ieeeconf}
\IEEEoverridecommandlockouts
\usepackage{iros}
\acrodef{qp}[QP]{quadratic programming}
\acrodef{fd}[FD]{force decomposition}
\acrodef{dof}[DoF]{degree-of-freedom}
\acrodef{ros}[ROS]{Robot Operating System}
\acrodef{uav}[UAV]{unmanned aerial vehicle}
\acrodef{uam}[UAM]{unmanned aerial manipulator}
\acrodef{com}[CoM]{center of mass}
\acrodef{mrr}[MRR]{modular reconfigurable robot}
\acrodef{vkc}[VKC]{virtual kinematic chain}

\title{\LARGE \bf Sequential Manipulation Planning for\\Over-actuated Unmanned Aerial Manipulators}

\author{Yao Su$^{1*}$, Jiarui Li$^{1,2*}$, Ziyuan Jiao$^{1*}$, Meng Wang$^{1}$, Chi Chu$^{1,3}$, Hang Li$^{1}$, Yixin Zhu$^{4}$, Hangxin Liu$^{1\dagger}$%
\thanks{$^{*}$ Equal contributors. $^\dagger$ Corresponding author.
$^{1}$ National Key Laboratory of General Artificial Intelligence, Beijing Institute for General Artificial Intelligence (BIGAI).
$^{2}$ Department of Advanced Manufacturing and Robotics, College of Engineering, Peking University.
$^{3}$ Department of Automation, Tsinghua University.
$^{4}$ Institute for Artificial Intelligence, Peking University.
Emails: \tt{\{suyao, lijiarui, jiaoziyuan, wangmeng, chuchi, lihang\}@bigai.ai}, \tt{yixin.zhu@pku.edu.cn}, \tt{liuhx@bigai.ai}%
}%
}

\begin{document}
\maketitle
\thispagestyle{empty}
\pagestyle{empty}

\begin{abstract}
We investigate the \textit{sequential} manipulation planning problem for \acp{uam}. Unlike prior work that primarily focuses on \textit{one-step} manipulation tasks, \textit{sequential} manipulations require coordinated motions of a \ac{uam}'s floating base, the manipulator, and the object being manipulated, entailing a unified kinematics and dynamics model for motion planning under designated constraints. By leveraging a \acf{vkc}-based motion planning framework that consolidates components' kinematics into one chain, the sequential manipulation task of a \ac{uam} can be planned as a whole, yielding more coordinated motions. Integrating the kinematics and dynamics models with a hierarchical control framework, we demonstrate, for the first time, an over-actuated \ac{uam} achieves a series of new sequential manipulation capabilities in both simulation and experiment. 
\end{abstract}

\vspace{6pt}

\section{Introduction}

Combining the agility of \acp{uav} and the flexibility of manipulators, \acp{uam} can conduct manipulation tasks across rough terrains and in regions unreachable by ground robots~\cite{fumagalli2014developing,ollero2021past,ryll2022smors}. Oftentimes, a fully- or even over-actuated \ac{uav} serves as the \acp{uam}' flying vehicle~\cite{jiang2017estimation,park2018odar,yi2020modeling}; this type of \acp{uav} can track position and orientation independently to provide the \ac{uam} with more agile maneuver, achieve a larger reachable workspace, and obtain better dynamic properties compared with traditional multirotors.
Existing \acp{uam} leverage a bi-level schema by combining (i) a controller to stabilize the system and track the desired trajectory under forceful contacts with the environment and (ii) a motion planner to produce trajectories satisfying task-related constraints. Such a bi-level schema has succeeded in various aerial manipulation tasks, such as pick-and-place~\cite{ryll2022smors,zhao2022versatile}, inspection~\cite{tognon2018control,bodie2021dynamic}, valve operation~\cite{zhao2022forceful}, and door-like articulated object manipulation~\cite{brunner2022planning,sugito2022aerial}.

Yet to date, \acp{uam} are limited to tasks with \textbf{one-step} planning. To endow with \textbf{multi-step} sequential manipulation capability, the \ac{uam} platform ought to (i) coordinate the motions of its floating base and the manipulator that consists of a series of revolute/prismatic joints, and (ii) effectively produce varied motion patterns at different steps of a sequential task, especially when interacting with objects with diverse kinematic structures. Developing such a sequential manipulation planning schema for \acp{uam} remains an unexplored topic.

\begin{figure}[t!]
    \centering   
    \includegraphics[width=\linewidth]{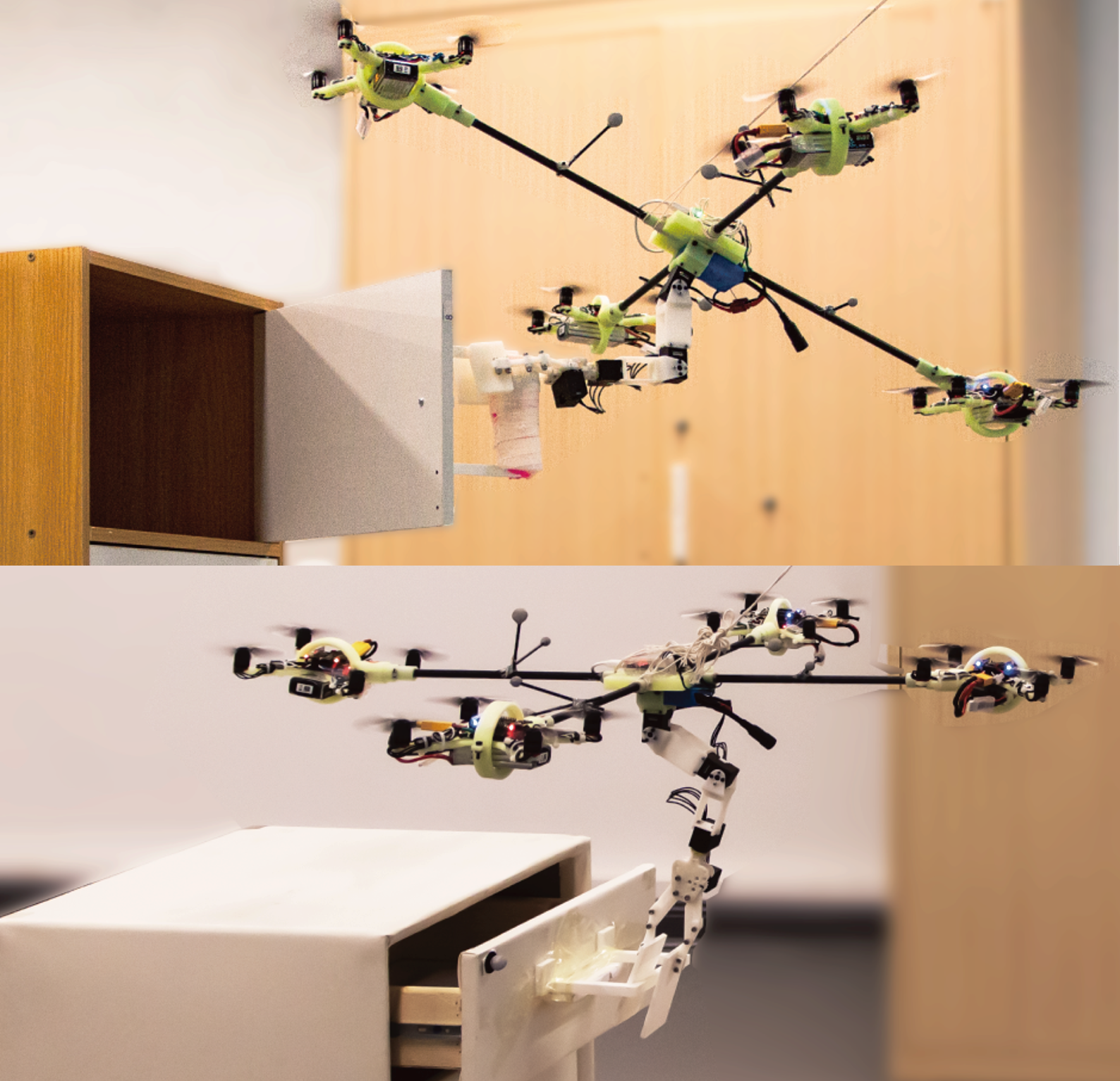}
    \caption{\textbf{Two sequential manipulation tasks completed by the \acs{uam}}. They require an over-actuated \acs{uam} platform for more agile motion, a \acs{vkc} modeling technique for manipulation planning, and an effective hierarchical control algorithm.}
    \label{fig:motive}
\end{figure}

\begin{figure*}[t!]
    \centering
    \includegraphics[width=\linewidth]{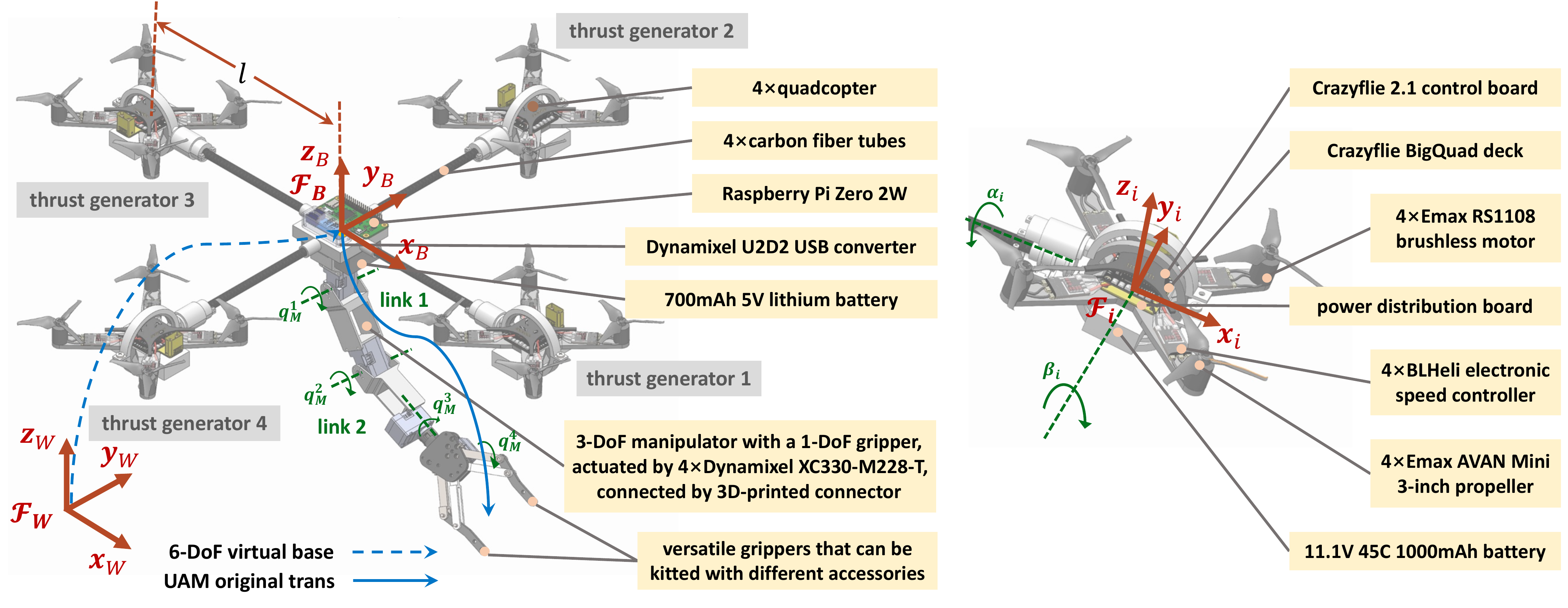}
    \caption{\textbf{Hardware design and attached coordinate frames of our over-actuated \acs{uam} platform.} The flying vehicle consists of four omnidirectional thrust generators. Each thrust generator has a 2-\acs{dof} passive gimbal mechanism and a quadcopter for over-actuation. The manipulator has three serial rotational \acsp{dof} and a parallel gripper.}
    \label{fig:platform}
\end{figure*}

Planning sequential manipulation is challenging even for ground mobile manipulators~\cite{berenson2008optimization,chitta2010planning,gochev2012planning}. In particular, consolidating the kinematics of the mobile base, the manipulator, and the manipulated object into one kinematics chain---constructing a \ac{vkc}---emerges as an effective means; it plans the mobile manipulator as a whole, yielding more coordinated manipulations~\cite{jiao2021consolidating,jiao2021efficient,haviland2022holistic,rofer2022kineverse}.

Inspired by \ac{vkc}, we extend the whole-body sequential manipulation from ground robots to \acp{uam}. Here, ``whole-body'' refers to the unification of the trajectory planning for the floating base and the motion planning for the manipulator.
First, we devise a novel \ac{uam}~\cite{su2021nullspace,su2022down,yu2021over} by integrating a 4-\ac{dof} manipulator with an over-actuated \ac{uav} that can be easily replicated by composing four modular quadcopters.
Next, through a dedicated nullspace-based control allocation framework, this new \ac{uam} platform possesses high thrust efficiency, can achieve arbitrary attitudes control, and is robust against controller sampling frequency and measurement noise~\cite{su2021nullspace,su2022down}.
Finally, after inserting \textit{virtual} linkages and joints and abstracting the object being manipulated by its kinematic structure, we derive the (virtual) kinematics and dynamics of this new \ac{uam} and solve the corresponding motion planning problems on the \ac{vkc} via trajectory optimization~\cite{jiao2021consolidating,jiao2021efficient}. 

\setstretch{0.99}

Our new \ac{uam} platform, integrated with \ac{vkc}-based planning framework and hierarchical control architecture, is demonstrated on various sequential aerial manipulation tasks involving multiple steps in simulations and physical experiments. \cref{fig:motive} depicts an example of relocating an object into a closed drawer and a closed cabinet, requiring six manipulation steps. This experiment, for the first time, demonstrates the plausibility and potential of planning sequential aerial manipulation tasks for \acp{uam}.

This paper makes three contributions: (i) We put forward a novel mechanical design of an over-actuated \ac{uam} and derive the dynamics model of the system. (ii) We devise a manipulation planning and hierarchical control framework for \acp{uam}. (iii) We demonstrate \acp{uam}' sequential manipulation capabilities in both simulations and experiments. 

The remainder of the paper is organized as follows. \cref{sec:hardware} presents the hardware design of our \ac{uam} platform. \cref{sec:model,sec:control,sec:planning} describe the system dynamics, manipulation planning, and control framework of this platform, respectively. Simulation and experimental results are summarized in \cref{sec:sim,sec:experiment}, respectively. \cref{sec:conclusion} concludes the paper.

\section{Hardware Design}\label{sec:hardware}

As shown in \cref{fig:platform}, our \ac{uam} platform consists of an over-actuated flying vehicle and a 4-\ac{dof} robotic manipulator connected at the bottom; it weights $1.21~kg$ with a maximum payload of $3~kg$. Due to limited onboard computing power, the platform's controller runs on a remote PC that wirelessly sends control commands to the platform.
 
\subsection{Flying Vehicle}

The flying vehicle's central frame is a rigid body made of a resin block fixed with four carbon-fiber tubes. Each tube is connected to an omnidirectional thrust generator with two added passive \acp{dof} to a generic quadcopter through a 3D-printed gimbal mechanism; see \cref{fig:platform}. Each quadcopter comprises a Crazyflie 2.1 control board, a Bitcraze's BigQuad Deck, and a power distribution board connected to an 11.1v Lithium battery. Four Emax RS1108 brushless motors actuate 3-inch propellers with a maximum thrust force of $t_{\text{max}}=2.6N$. Their speeds are controlled by an Electronic Speed Controller (ESC). This flying vehicle has been fully verified in prior work~\cite{su2022down,yu2021over,yu2023compensating}, capable of independently tracking 6-\ac{dof} position and attitude trajectory and achieving arbitrary attitude rotations with high thrust efficiency.

\subsection{Robotic Manipulator}

The robotic manipulator is installed at the bottom of the flying vehicle. It comprises three serial rotational \acp{dof} and a parallel gripper. Four Dynamixel XC330-M228-T motors are utilized to actuate the manipulator, and a Raspberry Pi Zero (RPi Zero) and a Dynamixel U2D2 converter are equipped on the flying vehicles to receive the control command wirelessly; RPi Zero send these signals to control the motors. The manipulator subsystem is powered by a $5~V$ battery. The system's physical properties are tabulated in \cref{tab:setup}.

\begin{table}[ht!]
    \centering
    \small
    \caption{\textbf{Physical parameters of the \ac{uam} platform.} $m_0$ and $I_0$ denote the mass and inertia matrix of the flying vehicle's mainframe, respectively. $m_i$ and $I_i$ denote each thrust generator's mass and inertia matrix, respectively. $m_M^j$ and $I_M^j$ denote the mass and inertia matrix of the manipulator link $j$, respectively.}
    \resizebox{0.85\linewidth}{!}{%
        \begin{tabular}{ccc}
            \toprule
            \textbf{Group} &  \textbf{Parameter} &  \textbf{Value}\\
            \midrule
            \multirow{6}{*}{flying vehicle}
                &  $m_B^0/kg$ & $0.168$\\
                &  $m_B^i/kg$ & $0.222$\\
                &  $\text{diag}(I_B^0)/kg \cdot cm^2$ & $[0.30\ 0.30 \ 0.60]$\\
                &  $\text{diag}(I_B^i)/kg \cdot cm^2$ & $[2.23\ 2.84\ 4.51]$\\
                &  $l/m$ & $0.21$\\
                &  $t_{\textit{max}}/N$ & $2.6$\\
            \midrule
            \multirow{7}{*}{manipulator}
                & $m_M^1/kg$   &$0.044$\\
                & $m_M^2/kg$   &$0.040$\\
                & $m_M^3/kg$   &$0.043$\\
                & $\text{diag}(I_M^1)/kg \cdot cm^2$   &$[0.22\ 0.21\ 0.04]$\\
                & $\text{diag}(I_M^2)/kg \cdot cm^2$   &$[0.22\ 0.19\ 0.06]$\\
                & $\text{diag}(I_M^3)/kg \cdot cm^2$   &$[0.82\ 0.80\ 0.15]$\\
                & griper range/$mm$   &$4-35$\\
            \midrule
            \multirow{4}{*}{others}
                & remote PC control rate/$Hz$ & $100$\\
                & quadcopter control rate/$Hz$ & $500$\\
                & manipulator control rate/$Hz$ & $500$\\
                & communication delay/$ms$ & $20$ \\
            \bottomrule
        \end{tabular}
     }%
    \label{tab:setup}
\end{table}

\section{Dynamics Modelling}\label{sec:model}

The complete dynamics model of the \ac{uam} platform is too complex for controller design; dividing it into two decoupled subsystems---the arm and the flying vehicle---introduces severe disturbance to the platform. As a result, we simplify the flying vehicle's dynamics by concentrating on compensating the gravity torque introduced by the shift of \ac{com} when the manipulator is in motion.

\subsection{Platform Configuration and Notation}

\cref{fig:platform} illustrates related coordination frames. The world frame and the \ac{uam}'s body frame are denoted as $\mathcal{F}_W$ and $\mathcal{F}_B$, respectively. We define the body frame's position as $\pmb{p}=[x,\, y, \,z]^\mathsf{T}$, the attitude in the roll-pitch-yaw convention as $\pmb{\theta}=[\phi,\,\theta,\,\psi]^\mathsf{T}$, and the angular velocity in $\mathcal{F}_B$ as $\pmb{\omega} = [p,\,q,\,r]^\mathsf{T}$~\cite{pi2021simple}. Frame $\mathcal{F}_{i}$ is attached to the $i$th thrust generator's center. We combine the flying vehicle's pose and velocity as $\pmb{q}_B=[\prescript{W}{}{\pmb{p}}^\mathsf{T},\,\prescript{B}{}{\pmb{\theta}}^\mathsf{T}]^\mathsf{T}$ and $\pmb{\dot{q}}_B=[\prescript{W}{}{\pmb{v}}^\mathsf{T},\,\prescript{B}{}{\pmb{\omega}} ^\mathsf{T}]^\mathsf{T}$. Let $\pmb{q}_M\in\mathbb{R}^{4\times1}$ be the manipulator's joint angles.

\subsection{Flying Vehicle Dynamics}

The dynamics model of the flying vehicle is simplified as:
\begin{equation}
    \resizebox{0.9\linewidth}{!}{%
        $\begin{bmatrix}
         m\pmb{I}_3 && 0\\
         0 && \prescript{B}{}{\pmb{J}}(\pmb{q}_M)
        \end{bmatrix}
        \pmb{\ddot{q}}_B
        = 
        \begin{bmatrix}
           \prescript{W}{B}{\pmb{R}}&0\\
            0& \pmb{I}_3
        \end{bmatrix}
        \pmb{u} + 
        \begin{bmatrix}
             mg \pmb{\hat{z}} \\
             \prescript{B}{}{\pmb{\tau}_g}(\pmb{q}_M) -\prescript{B}{}{\pmb{\omega}}\times \prescript{B}{}{\pmb{J}}(\pmb{q}_M)\prescript{B}{}{\pmb{\omega}}\\
        \end{bmatrix}$%
    },%
    \label{eq:quad_dyna}
\end{equation}
where $g$ is the gravitational acceleration, $m$ the whole platform's total mass, $\pmb{J}$ the whole platform's inertia matrix, $\prescript{B}{}{\pmb{\tau}_g}$ the gravitational torque due to the displacement of its \ac{com} from the geometric center, and $\hat{\pmb{z}} = \left[0,\,0,\,1\right]^\mathsf{T}$ the unit vector in the vertical direction in the world frame. Of note, $\pmb{J}$ and $\prescript{B}{}{\pmb{\tau}_g}$ are functions of the manipulator's joint angles $\pmb{q}_M$ defined by kinematic relationships. And
\begin{equation}
    \small
    \pmb{u} = 
    \begin{bmatrix}
        \displaystyle
        \sum_{i=1}^{4} \prescript{B}{i}{\pmb{R}} \,T_i \pmb{\hat{z}} \\
        \displaystyle
        \sum_{i=1}^{4} (\pmb{d}_i \times \prescript{B}{i}{\pmb{R}} \,T_i \pmb{\hat{z}}) \\
    \end{bmatrix}
    =    
    \begin{bmatrix}
        \pmb{J}_v(\pmb{\alpha},\pmb{\beta}) \\
        \pmb{J}_\omega(\pmb{\alpha},\pmb{\beta})
    \end{bmatrix}
    \pmb{T},
    \label{eq:bu_u}
\end{equation}
where $T_i$, $\alpha_i$, and $\beta_i$ denote the magnitude of thrust, tilting, and twisting angles of the $i$th thrust generator, respectively. $\pmb{d}_i$ is the distance vector from $\mathcal{F}_B$'s center to each $\mathcal{F}_{i}$.

\subsection{Manipulator Dynamics}

The dynamics of the manipulator are modeled following Luo \etal~\cite{luo2021modeling}, formally as:
\begin{equation}
   \small
   \pmb{M}_M(\pmb{q}_M)\pmb{\ddot{q}}_M+\pmb{C}_M(\pmb{q}_M,\pmb{\dot{q}}_M)+\pmb{G}_M(\pmb{q}_M)=\pmb{\tau}_M+\pmb{J}_{\textit{ext}}\,\pmb{F}_{\textit{ext}},
   \label{eq:manipulator_dyna}
\end{equation}
where $\pmb{M}_M\in\mathbb{R}^{4\times4}$ is the manipulator's inertia matrix, $\pmb{C}_M\in\mathbb{R}^{4\times1}$ the vector of the Coriolis and centrifugal terms, $\pmb{G}_M\in\mathbb{R}^{4\times1}$ the gravitational force vector, $\pmb{\tau}_M$ the torque command of each joint actuator, $\pmb{F}_{\textit{ext}}$ external forces, and $\pmb{J}_{\textit{ext}}$ the related Jacobian matrix.

\section{Sequential Aerial Manipulation Planning}\label{sec:planning}

In this section, we start by describing three essential steps to construct \acp{vkc}~\cite{jiao2021consolidating,jiao2021efficient,jiao2022planning} for our \ac{uam} platform. Next, we formulate the sequential manipulation planning problem on \acp{vkc} and solve it through trajectory optimization for aerial manipulation tasks.

\subsection{Modeling \texorpdfstring{\acp{uam}}{} with \texorpdfstring{\acp{vkc}}{}}

\textbf{Kinematic inversion} reverses the kinematic model of an articulated object by converting the attachable link into the new root in the inverted kinematic model. Of note, in addition to reversing the parent-child relationship for every two adjacent frames between the base link and the attachable link of the object, the spatial transformation of each joint must also be updated appropriately since a joint typically constrains the child link's motion \wrt child link's frame. 

\textbf{Virtual joint} defines the spatial transformation between two body frames and the joint type that constrains the relative motion between them. The manipulator and the manipulated object form a single serial kinematic chain by inserting a virtual joint between the manipulator's end-effector and an attachable link in the object model. If a manipulated object is articulate, its kinematic model has to be inverted for the constructed kinematic chain to remain serial.

\textbf{Virtual base} reflects the motion constraints imposed on the floating base. In our \ac{uam} platform, the floating base is an over-actuated \ac{uav} that can achieve free motion in space. Specifically, starting from the ground, we add three perpendicular prismatic joints for linear motion, followed by three revolute joints at the center of the \ac{uav} body frame for angular motions. These six joints together form a virtual chain that mimics the possible motions of the floating base.

A \ac{vkc} for aerial manipulation planning is constructed by augmenting a virtual base to the \ac{uam}'s kinematic model; see \cref{fig:platform}. During the manipulation, the end-effector connects to the inverted object model via a virtual joint. From this \ac{vkc} perspective, performing an aerial manipulation task is treated as altering the \ac{vkc}'s state, equivalent to solving a motion planning problem on \acp{vkc}.

\subsection{Motion Planning on \texorpdfstring{\acp{vkc}}{}}

The state vector $\pmb{x}\in\mathcal{X}_{\text{free}}$ describes the state of a \ac{vkc}, where $\mathcal{X}_{\text{free}}\in\mathbb{R}^n$ is the collision-free configuration space. The motion planning problem on \acp{vkc} is equivalent to finding a $T$-step path $\pmb{x}_{1:T}\in \mathcal{X}_{\text{free}}$, which can be formulated and solved by trajectory optimization.
Following Jiao \etal~\cite{jiao2021consolidating,jiao2021efficient}, the objective function of the trajectory optimization is:
\begin{equation}
    \small
    \underset{\pmb{x}_{1:T}}{\text{min}} \sum_{t=1}^{T-1} \norm{ \pmb{W}_{\text{v}}\delta{\pmb{x}}_{t} }_2^2
    \; + \sum_{t=2}^{T-1} \norm{ \pmb{W}_{\text{a}}\delta\dot{\pmb{x}}_{t} }_2^2,
    \label{eqn:objective}
\end{equation}
where we penalize the overall traveled distance and overall smoothness of the trajectory $\pmb{x}_{1:T}$. $\pmb{W}_{\text{v}}$ and $\pmb{W}_{\text{a}}$ are two diagonal weighting matrices for each \ac{dof}, $\delta{\pmb{x}}_{t}$ and $\delta\dot{\pmb{x}}_{t}$ are finite forward difference and second-order finite central difference of $\pmb{x}_{t}$, respectively. An equality constraint is imposed on the constructed \ac{vkc}, which specifies the physical constraints of the object or the environment: 
\begin{equation}
    \small
    h_{\text{chain}}(\pmb{x}_t) = 0, \; \forall t = 1, 2, \ldots, T.
    \label{eqn:chain_cnt}
\end{equation}
Failing to account for this type of constraint (\eg, the kinematic constraint of the manipulator or the object) may damage the \ac{uam} or the object being manipulated, resulting in failed executions.

The goal of a sequential aerial manipulation task is formulated as an inequality constraint:
\begin{equation}
    \small
    \norm{ f_{\text{task}}(\pmb{x}_T) - \mathcal{G}_{\text{goal}} }^2_2 \leq \xi_{\text{goal}}, \label{eqn:goal_cnt}
\end{equation}
which bounds the final state $\pmb{x}_T$ of the \ac{vkc} and the task goal $\mathcal{G}_{\text{goal}}\in\mathcal{G}$ with a small tolerance $\xi_{\text{goal}}$. The function $f_{\text{task}}: \mathbb{R}^n\rightarrow\mathbb{R}^k$ maps $\pmb{x}_T$ from the configuration space $\mathcal{X}$ to the task-dependent goal space $\mathcal{G}\in\mathbb{R}^k$. For example, in an object-picking task, $f_{\text{task}}$ represents the forward kinematics of the \ac{vkc}, and $\mathcal{G}_{\text{goal}}$ is the end-effector pose prior to grasping~\cite{jiao2022virtual}.

Additional safety constraints are further imposed on the motion planning problem:
\begin{align}
    \small
    &\pmb{x}_{\text{min}} \leq \pmb{x}_t \leq \pmb{x}_{\text{max}}, \, \;\;\; \forall t = 1, 2, \ldots, T%
    \label{eqn:joint_limit}%
    \\
    \norm{\delta{\pmb{x}}_{t}}_\infty \leq &\dot{\pmb{x}}_{\text{max}}, \norm{\delta\dot{\pmb{x}}_{t}}_\infty \leq \ddot{\pmb{x}}_{\text{max}}, \, \;\;\; \forall t = 2, 3, \ldots, T-1%
    \label{eqn:acc}%
    \\
    &\sum_{i= 1}^{N_{\text{link}}} \sum_{j=1}^{N_{\text{obj}}} |\text{dist}_{\text{safe}} - f_{\text{dist}}(L_i, O_j)|^{+} \leq \xi_{\text{dist}}%
    \label{eqn:collision_1},%
    \\
    &\sum_{i= 1}^{N_{\text{link}}} \sum_{j=1}^{N_{\text{link}}} |\text{dist}_{\text{safe}} - f_{\text{dist}}(L_i, L_j)|^{+} \leq \xi_{\text{dist}}%
    \label{eqn:collision_2},%
\end{align}
where $|\cdot|^{+}$ is defined as $|x|^{+}=\text{max}(x,0)$.
\cref{eqn:joint_limit,eqn:acc} are inequality constraints that define the joint capability and implicitly constrain the workspace of a \ac{uam}. \cref{eqn:collision_1,eqn:collision_2} penalize collisions with obstacles and self-collisions, respectively. $\text{dist}_{\text{safe}}$ is a predefined safety distance, and $f_{\text{dist}}$ is a function that calculates the signed distance between a pair of objects.

\section{Control}\label{sec:control}

\begin{figure}[b!]
    \centering
    \includegraphics[width=\linewidth]{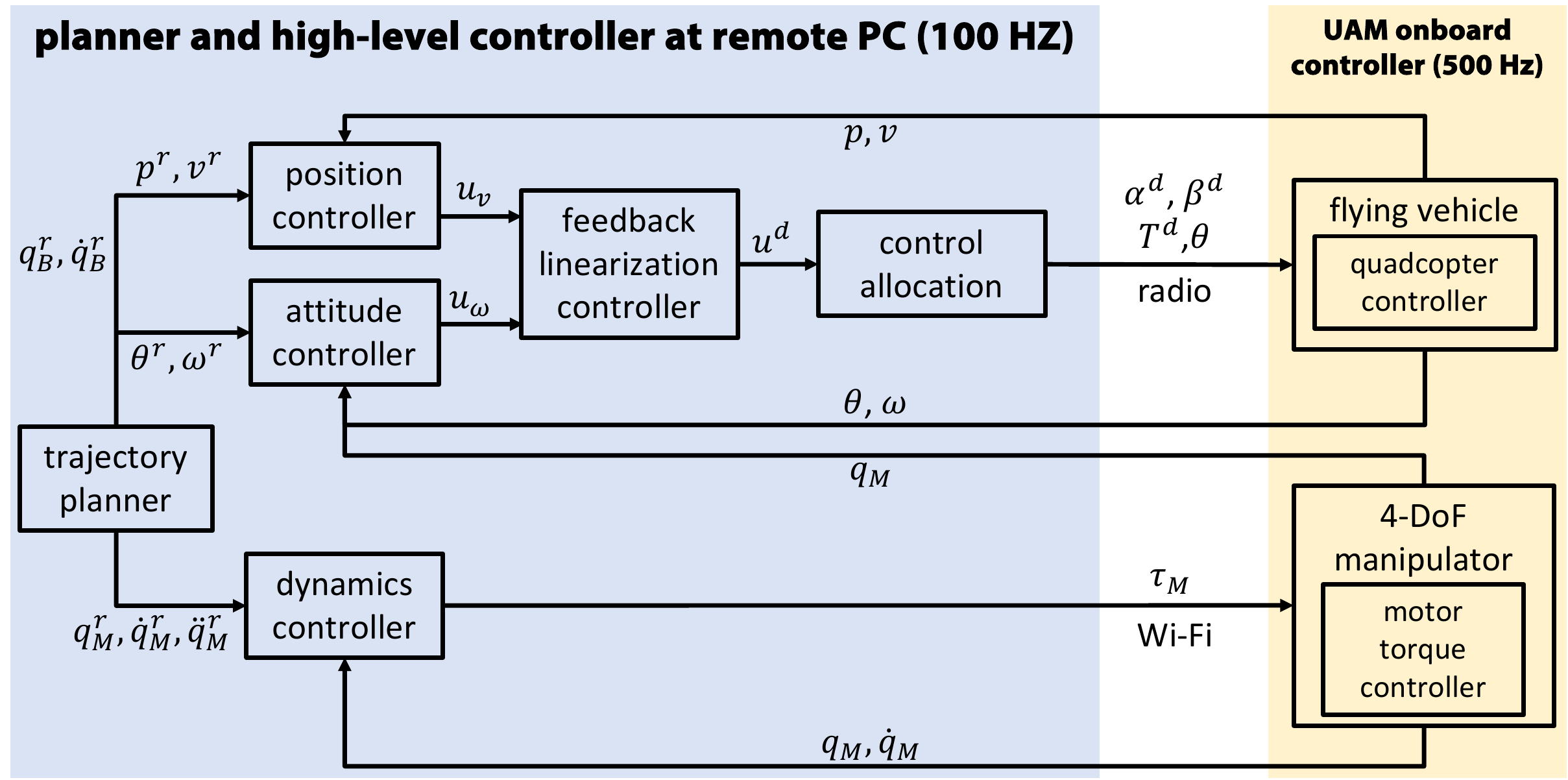}
    \caption{\textbf{Hierarchical control architecture of the \ac{uam} platform}. The high-level controller of the flying vehicle (i) calculates desired wrench command $\pmb{u}^d$ for trajectory tracking and (ii) allocates it to desired thrusts and joint angles of thrust generators through control allocation. Each quadcopter has its own onboard controller to regulate the joint angles and thrust to desired values.}
    \label{fig:control}
\end{figure}

Using a hierarchical control architecture, we devise the \ac{uam}'s overall controller with two subsystems, shown in \cref{fig:control}. The high-level controller calculates the desired commands for trajectory tracking remotely and sends them wirelessly to the low-level controller that runs on the platform with high frequency.

\subsection{Flying Vehicle Control}

\paragraph*{High-level control}

The feedback-linearization method is applied to \cref{eq:quad_dyna} to transfer the nonlinear system dynamics to a linear double integrator~\cite{ruan2023control,su2021fast,su2023fault}:
\begin{equation}
\small
        \pmb{u}^d=
            \begin{bmatrix}
                \displaystyle
                {m}\prescript{W}{B}{\pmb{R}^\mathsf{T}}\left(\pmb{u}_{v}-g \pmb{\hat{z}}\right)
                \\
                \displaystyle
                 \prescript{B}{}{\pmb{J}}(\pmb{q}_M)\pmb{u}_{\omega}-\left(\prescript{B}{}{\pmb{\tau}_g}(\pmb{q}_M)-\prescript{B}{}{\pmb{\omega}}\times \prescript{B}{}{\pmb{J}}(\pmb{q}_M)\prescript{B}{}{\pmb{\omega}}\right)
            \end{bmatrix}
    \label{eq:ud}
\end{equation} 
where the superscript $d$ indicates the desired values, $\pmb{u}_v$ and $\pmb{u}_\omega$ are two virtual inputs that can be designed with translational and rotational errors to track the reference position and attitude trajectory:
\begin{equation}
    \small
    \begin{aligned}
        \pmb{u}_{v}&= \pmb{\dot{{v}}}^r + K_{{v}1}\pmb{e}_{v}+ K_{{v}2}\pmb{e}_{p}+K_{{v}3}\int{\pmb{e}_{p}}\,dt,\\
        \pmb{u}_{\omega}&= \pmb{\dot{\omega}}^r + K_{\omega1}\pmb{e}_\omega+ K_{\omega2}\pmb{e}_\theta+K_{\omega3}\int{\pmb{e}_\theta}\,dt,
        \label{eq:pos_and_att_control}
    \end{aligned}   
\end{equation}
where $K_{p i}$ and $K_{\omega i}$ are constant gain matrices, and the superscript $r$ indicates the reference value from the \ac{vkc}-based motion planning; see \cref{sec:planning}. The error terms are defined following Su \etal~\cite{su2021compensation}:
\begin{equation}
    \small
    \begin{aligned}
        \pmb{e}_{p} &= \pmb{p}^r - \pmb{p}, \quad \pmb{e}_{v} = \pmb{v}^r - \pmb{v},\\ 
        \pmb{e}_{\theta}&=\frac{1}{2}[\pmb{R}(\pmb{\theta})^T\pmb{R}(\pmb{\theta}^r)-\pmb{R}(\pmb{\theta}^r)^T\pmb{R}(\pmb{\theta})]_\vee, \\
        {\pmb{e}}_\omega  &=\pmb{R}(\pmb{\theta})^T\pmb{R}(\pmb{\theta}^r)\pmb{\omega}^r - \pmb{\omega},
        \label{eq:pos_and_att_err}
    \end{aligned}   
\end{equation}
where $\pmb{R}\left(\cdot\right)$ is the transformation from Euler angles to a standard rotation matrix, and $\left[\cdot\right]_\vee$ is the mapping from SO(3) to $\mathbb{R}^3$. Combining \cref{eq:ud,eq:pos_and_att_control,eq:pos_and_att_err} with \cref{eq:quad_dyna}, we have the error dynamics as: 
\begin{equation}
    \small
    \begin{aligned}
        \dot{\pmb{e}}_{v}+ K_{{v}1}\pmb{e}_{v}+ K_{{v}2}\pmb{e}_{p}+K_{{v}3}\int{\pmb{e}_{p}}\,dt&=0,\\
        \dot{\pmb{e}}_{\omega}+ K_{\omega1}\pmb{e}_\omega+ K_{\omega2}\pmb{e}_\theta+K_{\omega3}\int{\pmb{e}_\theta}\,dt&=0,
    \end{aligned}   
\end{equation}
which is an asymptotically stable system.

\paragraph*{Control allocation and low-level control}

The control allocation solves for desired command $\alpha_i^d$, $\beta_i^d$, and $T_i^d$ for each 3-\ac{dof} thrust generator from total wrench command of whole flying vehicle $\pmb{u}^d$. Among various approaches~\cite{yu2021over,ruan2023control,su2021nullspace,su2022down,ryll2014novel}, we implement the downwash-aware control allocation method~\cite{su2022down} to avoid the large disturbance caused by downwash flows that counteract other thrust generators while maintaining high thrust efficiency, critical for a smooth aerial manipulation, especially when interacting with an object.

In low-level control, two separated PID controllers are designed to allow each quadcopter to track the desired tilting and twisting angles, $\alpha_i^d$ and $\beta_i^d$, with tilting torque commands. This is combined later with the thrust force command $T_i^d$ to determine each actuator's angular velocity. Finally, it is converted to a PWM command to drive the actuators~\cite{yu2021over}.

\subsection{Manipulator Control}

With \cref{eq:manipulator_dyna}, we design the manipulator's controller,
\begin{equation}
    \small
    \pmb{\tau}_M=\pmb{M}_M(\pmb{q}_M)\pmb{\ddot{q}}_M^d+\pmb{C}_M(\pmb{q}_M,\pmb{\dot{q}}_M)+\pmb{G}_M(\pmb{q}_M),
    \label{eq:manipulator_control}
\end{equation}
where
\begin{equation}
    \small
    \begin{aligned}
        \pmb{\ddot{q}}_M^d&=\pmb{\ddot{q}}_M^r+K_{M1}\pmb{e}_M+K_{M2}\dot{\pmb{e}}_M+K_{M3}\int{\pmb{e}_M}\,dt, \\ 
        \pmb{e}_M&=\pmb{q}_M^r-\pmb{q}_M,
    \end{aligned}
\end{equation}
where $K_{M i}$ are constant gain matrices.

\section{Simulation}\label{sec:sim}

\subsection{Setup}

Before conducting experiments, we developed a simulation platform in MATLAB Simulink/Simscape to evaluate the proposed sequential manipulation planning framework on our customized over-actuated \ac{uam} platform~\cite{su2024flight,su2024marvel}. To realistically replicate the physical system, the simulator incorporates the \ac{uam}'s physical parameters, the dynamics of propeller motors and saturation, control frequencies, communication noise, measurement noise, and delays; see \cref{tab:setup}.

\begin{figure}[t!]
    \centering  
    \begin{subfigure}[b]{\linewidth}
        \includegraphics[width=\linewidth,trim=2cm 5.5cm 6cm 3cm, clip]{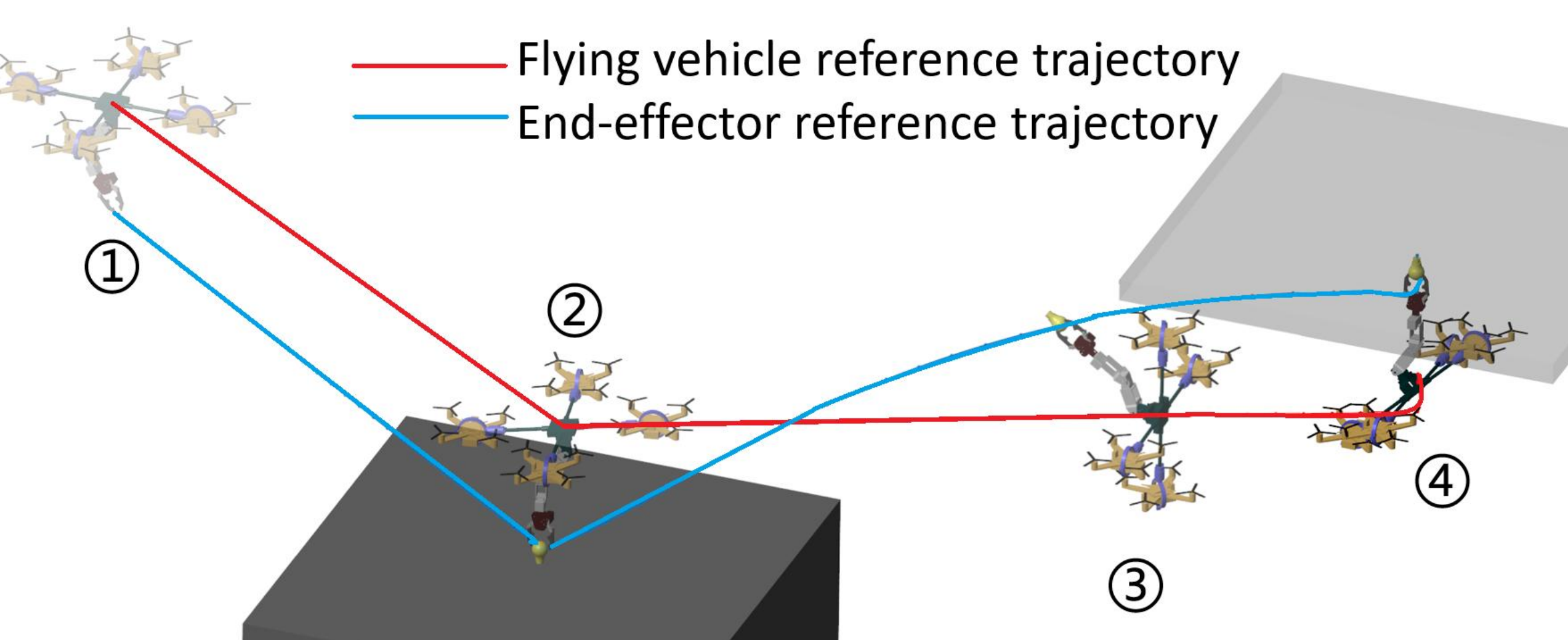}
        \caption{Keyframes in simulation}
        \label{fig:sim1_env}
    \end{subfigure}%
    \\%
    \begin{subfigure}[b]{\linewidth}
        \centering\includegraphics[width=\linewidth,trim=0cm 5cm 6.4cm 2.5cm, clip]{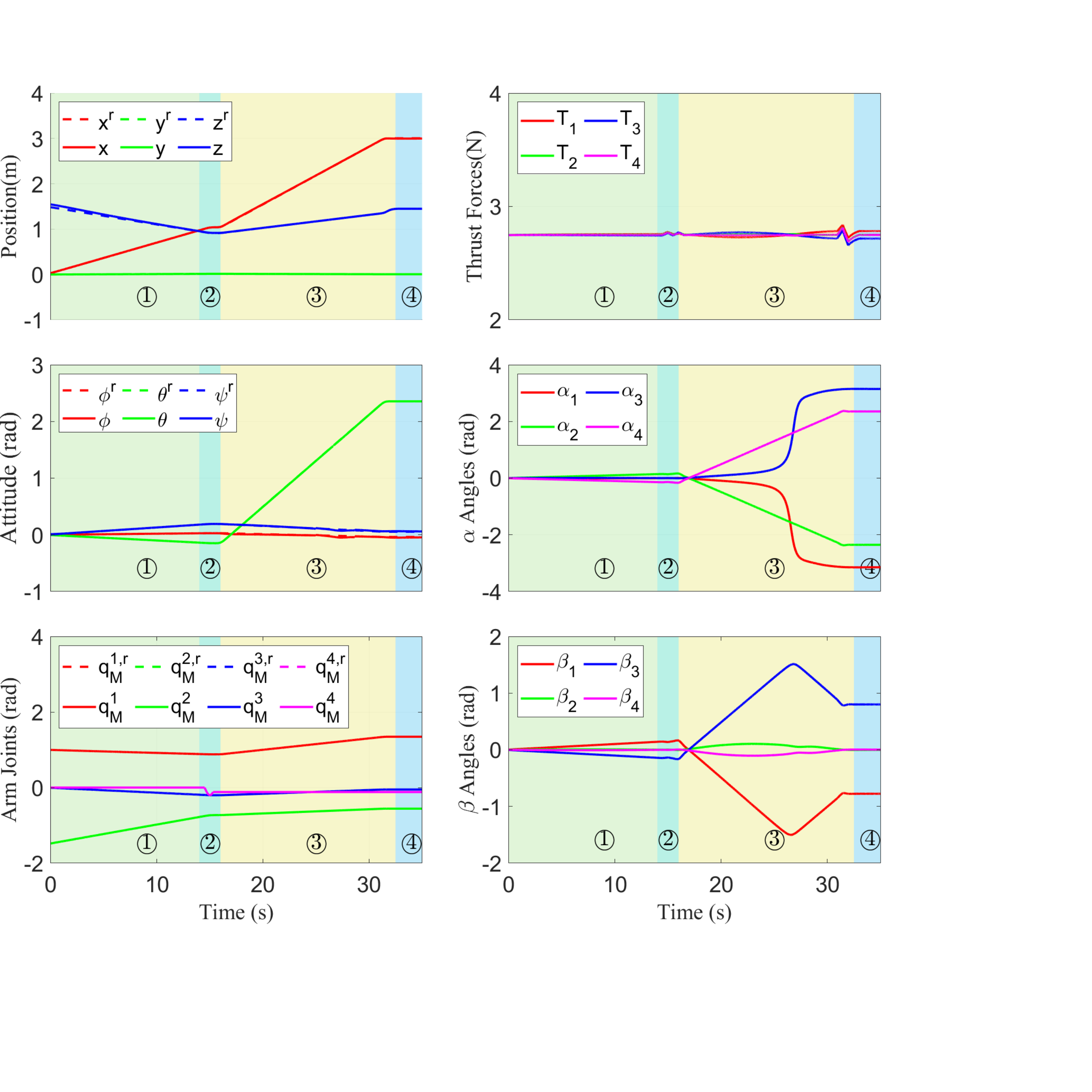}
        \caption{Simulation results}
        \label{fig:sim1_results}
    \end{subfigure}%
    \caption{\textbf{Simulation Task 1: Install a light bulb.} The action sequence of the \ac{uam} can be divided into four steps: \ding{172} approach, \ding{173} pick-up, \ding{174} rotate and translate, and \ding{175} feed in.}%
    \label{fig:sim1}
\end{figure}

\begin{figure}[t!]
    \centering  
    \begin{subfigure}[b]{\linewidth}
        \includegraphics[width=\linewidth,trim=0cm 0cm 11cm 0cm, clip]{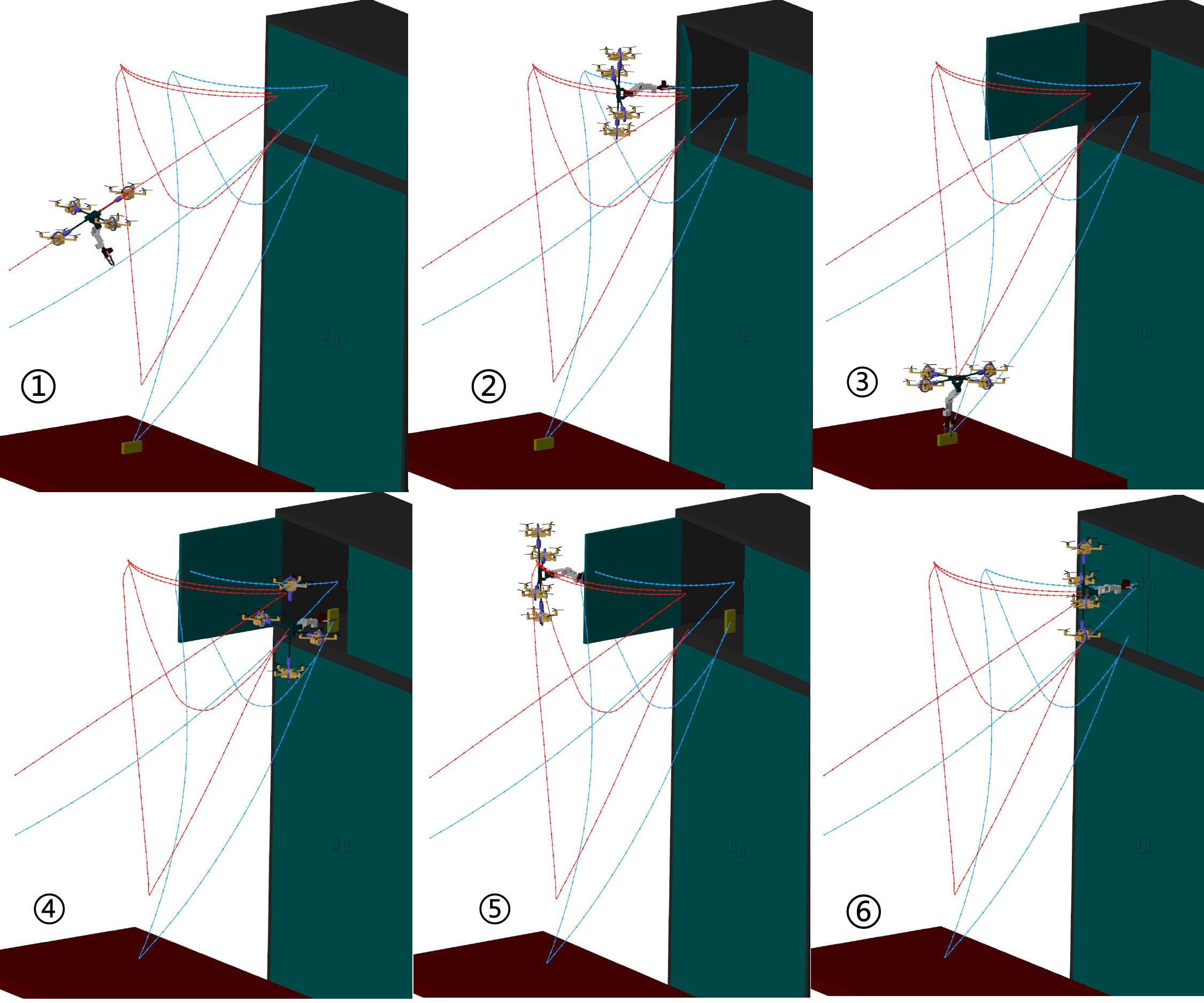}
        \caption{Keyframes in simulation}
        \label{fig:sim2_env}
    \end{subfigure}%
    \\%
    \begin{subfigure}[b]{\linewidth}
        \centering\includegraphics[width=\linewidth,trim=0cm 7.8cm 0cm 2cm, clip]{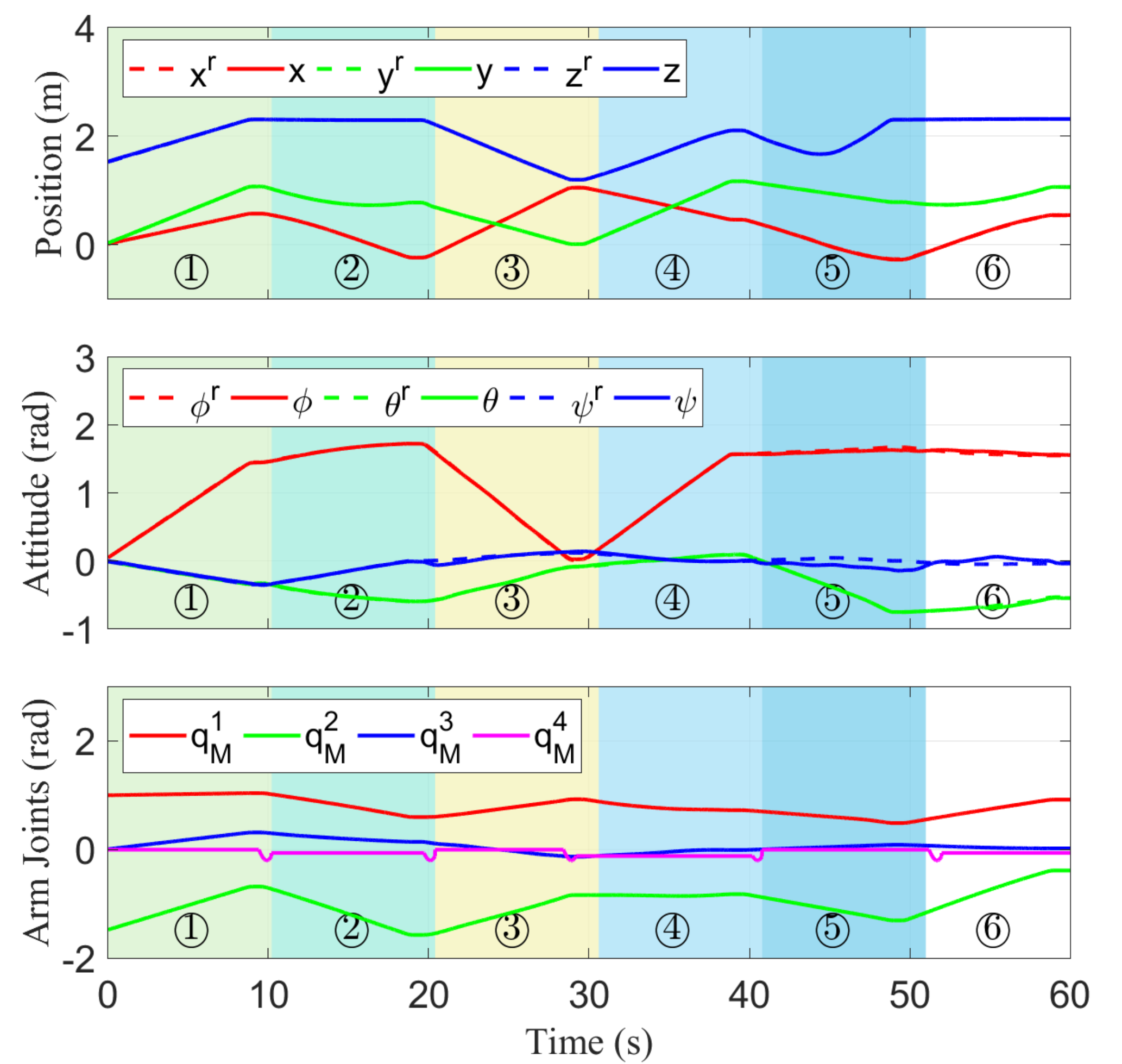}
        \caption{Simulation results}
        \label{fig:sim2_results}
    \end{subfigure}%
    \caption{\textbf{Simulation Task 2: Relocate an object into a cabinet.} The action sequence of the \ac{uam} are divided into six steps: \ding{172} approach to the door, \ding{173} open the door, \ding{174} pick up the object, \ding{175} put the object into the cabinet, \ding{176} approach to the door, and \ding{177} close the door.}%
    \label{fig:sim2}
\end{figure}

 \subsection{Results} 

\cref{fig:sim1,fig:sim2} summarize the simulation results of accomplishing two sequential aerial manipulation tasks---installing a light bulb and relocating an object into a closed cabinet---using the proposed manipulation planning framework.

\paragraph*{Task 1}

As shown in \cref{fig:sim1_env}, the light bulb installation task was divided into four steps in our \ac{vkc}-based motion planning framework: \ding{172} approach the light bulb, \ding{173} pick it up with the manipulator, \ding{174} flip the platform to transport the light bulb to the bottom of the target position, and \ding{175} move up to install it. A 10-\ac{dof} \ac{vkc} for the \ac{uam} platform is built---six for the flying vehicle and four for the manipulator. A feasible and collision-free reference trajectory is acquired within the physical constraints. As shown in \cref{fig:sim1_results}, with the hierarchical controller introduced in \cref{sec:control}, the proposed \ac{uam} platform can accurately track the reference trajectory to accomplish the task.

\paragraph*{Task 2}

Relocating an object into the cabinet requires a six-step action sequence of the \ac{uam}: \ding{172} approach the door and grasp the handle with the manipulator, \ding{173} open the door, \ding{174} ungrasp the handle and move to pick up the object, \ding{175} put the object into the cabinet, \ding{176} approach and grasp to the handle again, and \ding{177} close the door. Some keyframes are shown in \cref{fig:sim2}, and the planned reference trajectory by the \ac{vkc}-based motion planner and the tracking performance are shown in \cref{fig:sim2_results}.

These simulation results indicate that the \ac{vkc}-based motion planning framework and the proposed \ac{uam} platform effectively achieve sequential aerial manipulation.

\section{Experiment}\label{sec:experiment}

\subsection{Setup}

To further demonstrate the sequential aerial manipulation capability, we conduct experiments by implementing a table arrangement task in the physical world. Specifically, we use the Vicon motion capture system (MoCap) to measure the position and attitude of the \ac{uam} platform. The trajectory planner and main controller of the \ac{uam} systems runs on a remote PC (AMD Ryzen9 5950X CPU, 64 GB RAM), which communicates with the MoCap through Ethernet and efficiently solves the controller commands. The flying vehicle's primary controller is modified from the Crazyflie python library; it calculates the desired thrust $\pmb{T}^d$, tilting angles $\pmb{\alpha}^d$, and twisting angles $\pmb{\beta}^d$ for all quadcopter modules of the thrust generators and sends them through Crazy Radio PA antennas (2.4G~$Hz$). Each quadcopter is embedded with an onboard IMU module. Its firmware is modified to estimate the rotation angle given the attitude of central frame $\pmb{\theta}$, regulates the tilting and twisting angles to desired values with two PID loops, and provides the required thrust with 500~$Hz$ for a fast low-level response.

The manipulator controller is modified from the Robotis Dynamixel SDK, which runs on the RPi Zero. It wirelessly receives commands from the remote PC and sends commands to the motors through the Dynamixel U2D2 converter. Each Dynamixel XC330-M228-T motor has its own controller, and the control mode is set as current control. The measurement rate of the motion capture system, the remote PC controller, and the data communication with the \ac{uam} platform is all set to 100~$Hz$.

\subsection{Results}

\begin{figure}[t!]
    \centering 
    \begin{subfigure}[b]{\linewidth}
        \includegraphics[width=\linewidth,trim=0cm 6cm 1cm 2cm, clip]{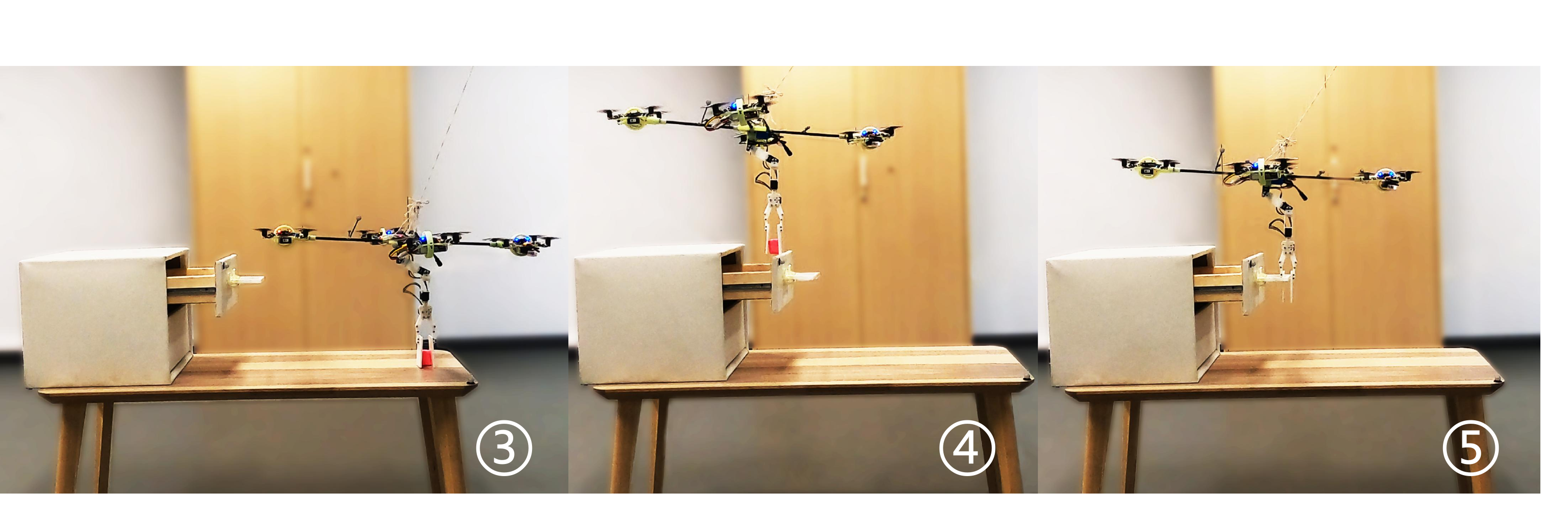}
        \caption{Keyframes in experiment}
        \label{fig:exp_clip}
    \end{subfigure}
    \\
    \begin{subfigure}[b]{\linewidth}
        \includegraphics[width=\linewidth,trim=0.6cm 0.8cm 0.5cm 0cm, clip]{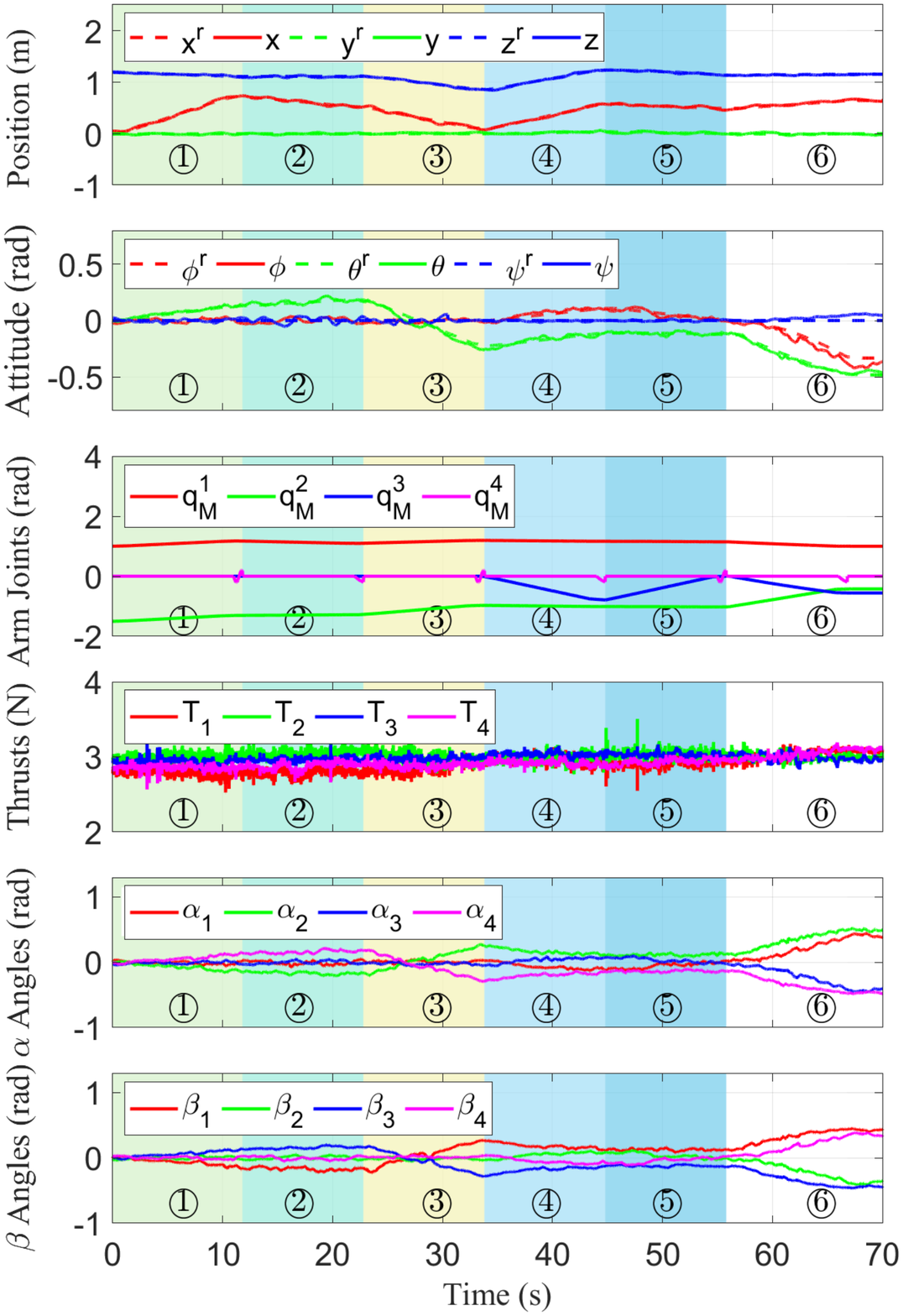}
        \caption{Experiment results}
        \label{fig:exp_results}
    \end{subfigure}    
    \caption{\textbf{The task for experiments: Relocate an object into the drawer.} The action sequence of the \ac{uam} is divided into six
    steps: \ding{172} approach to the drawer, \ding{173} open the drawer, \ding{174} approach the toy and pick it up, \ding{175} drop off the toy to the drawer,  \ding{176} approach to the drawer handle, and \ding{177} close the drawer.}
    \label{fig:exp}
\end{figure}

\cref{fig:exp} summarizes the experimental results of relocating an object into the drawer, which was divided into six steps: \ding{172} approach to the drawer and grasp the handle with the manipulator, \ding{173} open the drawer, \ding{174} ungrasp with the handle and move to pick up the toy, \ding{175} move to the opened drawer and drop off the toy inside, \ding{176} approach and grasp the handle again, and \ding{177} close the drawer. The 10-\ac{dof} reference trajectory provided by our \ac{vkc}-based motion planner was accurately tracked by the \ac{uam} with the hierarchical controller, and the desired commands for each omnidirectional thrust generator are plotted in \cref{fig:exp_results}. \cref{fig:motive,fig:exp_clip} show some experiment keyframes.

\section{Conclusion}\label{sec:conclusion}

In this paper, we presented a solution to the sequential aerial manipulation problem of \acp{uam}, an unexplored topic until now. Unlike previous work in \ac{uam} that solves the motion planning and control problems of one-step manipulation tasks, accomplishing sequential aerial manipulation requires (i) a highly efficient \ac{uam} platform, (ii) a specialized motion planner that can well-coordinate motions of the flying vehicle, the manipulator, and the manipulated object under varied settings, and (iii) an effective control scheme to track the desired trajectory. To jointly tackle these challenges, we designed a novel \ac{uam} platform based on an over-actuated \ac{uav} that can achieve omnidirectional flight with high thrust efficiency. To produce a long sequence of motions that coordinates well with each other, we extended the idea of \ac{vkc} used for ground mobile manipulators and developed a \ac{vkc}-based aerial manipulation planning framework for \acp{uam}. Together with a hierarchical control scheme, we validated our solution in both simulation and experiment. The results demonstrated that our approach endowed a new capability of sequential aerial manipulation for \acp{uam} and could open up new venues in the field of aerial manipulation.

The integration of a wireless tactile senor~\cite{li2023l3ftouch,liu2017glove} with the manipulator will be the future work for our research, which extends the manipulation capability of our \ac{uam} platform and enlarges the application range of our the \ac{vkc}-based planning framework to more complicated sequential aerial manipulation tasks.

\textbf{Acknowledgement:}
We thank Dr. Zeyu Zhang (BIGAI), Zhen Chen (BIGAI), Yangyang Wu (BIGAI), Zihang Zhao (BIGAI), Hao Liang (BIGAI), and Qing Lei (PKU) for their help on Vicon, figures, and hardware design. This work is supported in part by the National Key R\&D Program of China (2021ZD0150200) and the Beijing Nova Program.

\balance
\bibliographystyle{ieeetr}
\bibliography{reference_header,reference}

\end{document}